Artificial Intelligence versus Maya Angelou: Experimental evidence that people cannot differentiate AI-generated from human-written poetry

Nils Köbis[1, 2]* & Luca D. Mossink[1]

[1] Department of Economics & Center for Experimental Economics and Political Decision making (CREED), University of Amsterdam

[2] Max Planck Institute for Human Development, Center for Humans & Machines

* **Correspondence**: Nils Köbis (n.c.kobis@gmail.com), Center for Research in Experimental Economics and Political Decision Making, University of Amsterdam, Amsterdam, The Netherlands.

**Funding**: This project has received funding from the European Research Council (ERC) under the European Union's Horizon 2020 research and innovation programme (grant agreements: ERC-StG-637915; ERC-AdG 295707) from Research Priority Area Behavioral Economics (University of Amsterdam, proposal number 201906250406).

**Conflict of Interest:** The authors declare that they do not have any conflict of interests.

**Acknowledgements**: We thank Bobbie Goettler, Silvia Dominguez Martinez, Margarita Leib, Giulia Ranzini, Shaul Shalvi, Ivan Soraperra, Christopher Starke, Koen van der Veen, and the participants of the CREED Lunch Seminar for useful comments, suggestions and assistance.

**Abstract**

The release of openly available, robust natural language generation algorithms (NLG) has spurred much public attention and debate. One reason lies in the algorithms' purported ability to generate human-like text across various domains. Empirical evidence using incentivized tasks to assess whether people (a) can distinguish and (b) prefer algorithm-generated versus human-written text is lacking. We conducted two experiments assessing behavioral reactions to the state-of-the-art Natural Language Generation algorithm GPT-2 ($N_{total}$ = 830). Using the identical starting lines of human poems, GPT-2 produced samples of poems. From these samples, either a random poem was chosen (*Human-out-of-the-loop*) or the best one was selected (*Human-in-the-loop*) and in turn matched with a human-written poem. In a new incentivized version of the Turing Test, participants failed to reliably detect the algorithmically-generated poems in the *Human-in-the-loop* treatment, yet succeeded in the *Human-out-of-the-loop* treatment. Further, people reveal a slight aversion to algorithm-generated poetry, independent on whether participants were informed about the algorithmic origin of the poem (*Transparency*) or not (*Opacity*). We discuss what these results convey about the performance of NLG algorithms to produce human-like text and propose methodologies to study such learning algorithms in human-agent experimental settings.

**Artificial Intelligence versus Maya Angelou: Experimental evidence that people cannot**

**differentiate AI-generated from human-written poetry**

Artificial intelligence (AI), "the development of machines capable of sophisticated (intelligent) information processing" (Dafoe, 2018, p. 5), is rapidly advancing and has begun to take over tasks previously performed solely by humans (Rahwan et al., 2019). Algorithms are already assisting humans in writing text, such as autocompleting sentences in emails and even helping writers write novels (Streitfeld, 2018). Besides supporting humans, such natural-language generation (NLG) algorithms can autonomously create different types of texts. Already in use in the field of digital journalism, algorithms can generate news pieces based on standardized input data, such as sports scores or stock market values (van Dalen, 2012). However, autonomous *creative*-text generation presents a bigger challenge because it requires the creation of original content that is deemed appealing and useful (Bakhshi, Frey, & Osborne, 2015). Hence, creative writing has long been considered an impregnable task for algorithms (Keith, 2012; Penrose, 1990).

Yet, more recent developments in machine learning have expanded the scope and capacities of NLG (Jozefowicz, Vinyals, Schuster, Shazeer, & Wu, 2016). A notable case is the open-source algorithm called Generative Pre-Training 2 (GPT-2; Radford et al., 2019). At close to zero marginal cost, it produces text across a wide variety of domains, ranging from non-fiction, such as news pieces, to fiction, such as novels. The text outputs adhere to grammatical and semantical rules, and allegedly reach human levels. Due to such claims about the unprecedented abilities and the potential ethical challenges it raises, for example, as a tool for disinformation (Kreps & McCain, 2019), much controversy accompanied the algorithm's release (*The Guardian*, 2019).

However, systematic empirical examination of these claims is largely lacking – especially from an experimental social science perspective. In particular, whether humans are able to reliably distinguish creative text that is generated by an algorithm from one written by a human, when they are incentivized to do so, remains unknown. Do people prefer creative text written by fellow humans to that generated by algorithms? Does the information about the respective origin – being a human or an algorithm – sway this preference for the creative-text output? Does human involvement (or lack of) in the selection of the text output generated by the algorithm make a difference? To address these questions, we use incentivized paradigms to extend previous research into computational creativity by emphasizing the *human behavioral reactions* to NLG. Although much research has focused on the machinery – how to design algorithms to write creative text (Boden, 2009; Gonçalo Oliveira, 2018; Oliveira, 2009) – research on humans' behavioral reactions to such algorithms is much less developed.

**1.1. Distinguishing between Artificial and Human Text**

To gain empirical insights into people's ability to discern artificial from human content, we draw on the famous Turing Test (Saygin, Chaminade, Ishiguro, Driver, & Frith, 2012; Turing, 1950). Based on a thought experiment, Turing proposed it as a way to find out whether machines can think. The basic set-up entails three agents: one judge and two participants. The judge seeks to find out which of the other two participants is a machine and which is a human. In the classic version, the judge has five minutes to ask multiple questions and receives answers from the participants, after which the judge indicates which of the two is a human and which is a machine.

Since its introduction, various algorithms have attempted to pass the test in multiple tournaments and competitions (for an overview, see Warwick & Shah, 2016). In 2014, a chat bot, called Eugene Goostman, was claimed to have passed the Turing Test, by tricking 33% of human judges into believing they were communicating with a 13-year-old Ukrainian boy (Marcus, Rossi, & Veloso, 2016; Walsh, 2017; Warwick & Shah, 2016). Hence, a deceptive strategy of pretending to have rudimentary English proficiency paid off. Therefore, many scholars have criticized the standard Turing Test for identifying deceptive ability rather than intelligence (e.g., see Riedl, 2014; Walsh, 2017). In pursuit of better measures of machine intelligence, many extensions, modifications, and alternative tests have been proposed (for an overview, see the special issue on the subject in *AI Magazine*, Marcus et al., 2016).

However, according to the results of a systematic literature review (for more details, see SOM), no version of the Turing Test has contained financial incentives for judges' accuracy. That is, judges typically do not receive any financial reward for successfully detecting the human among the competitors. Hence, whether people are unable, or might simply not be motivated, to differentiate human from machine counterparts remains somewhat unknown. Filling this gap, we introduce a new version of the Turing Test that entails incentives for judges' accuracy. In both studies, we tested the prediction that humans' accuracy in correctly identifying whether the text is human-written or algorithm-generated does not exceed random guessing.

## 1.2. Overconfidence about Algorithmic Detection

Besides examining people's ability to detect human-written from algorithm-generated text, it is crucial to understand whether people have accurate beliefs about their own ability in that domain. Multiple studies have revealed overconfidence, and hence the systematic overestimation of one's own capabilities (Kruger & Dunning, 1999). While generally causing personal and social harm (Malmendier & Tate, 2008; Moore & Healy, 2008), overconfidence in the domain of algorithm detection poses the threat of making people especially vulnerable to deception. That is, when people overestimate their own abilities to detect algorithmic behavior, and fail to reliably do so, they can fall prey to being exposed to, and potentially influenced by, algorithms without noticing. To examine whether the commonly observed phenomenon of overconfidence also exists for algorithmic detection, we tested the hypothesis that people's perceived ability to detect algorithms systematically exceeds their actual accuracy levels.

## 1.3. Artificial Creativity: Aversion and Appreciation

Classically, machines have been seen as static rule-based systems. Because creativity requires the generation of original and useful ideas (Amabile, 1983), the possibility that machines could be creative was historically considered unfathomable. In fact, creativity still provides a big obstacle for machines that merely rely on automation (Bakhshi et al., 2015). Yet, recent advances in machine learning (ML) have increasingly enabled computers to "'learn' and change their behaviour through search, optimisation, analysis or interaction, allowing them to discover new knowledge or create artefacts which exceed that of their human designers in specific contexts" (McCormack & D'Inverno, 2014, p. 2). Hence, algorithms are becoming increasingly able to adapt, learn, and create original, unpredictable outputs.

ML has also changed the field of computational creativity (Boden, 2009; Loller-Andersen & Gambäck, 2018; Majid al-Rifaie, Cropley, Cropley, & Bishop, 2016; Oliveira, 2009; Sloman, 2012). Multiple algorithms have been developed to serve single creative purposes such as generating story narratives (Bringsjord & Ferrucci, 1999), crafting jokes (Ritchie et al., 2007), or writing poetry (for an overview, see Oliveira, 2009). Although these algorithms have been programmed with single purposes – for example, creating poetry – recent developments in transfer learning have rendered algorithms capable of text generation across various domains. The algorithm GPT-2, released in 2019 by OpenAI, is one of the most famous examples of such a robust text-generating algorithm. In short, using ML technology, GPT-2 is a transformer-based language model, trained on an unprecedentedly large dataset, to predict the next word for a given textual input (for more details on the algorithm, see Radford et al., 2019). Due to these extensive training datasets, the algorithm has a more thorough ability to reproduce syntax and thus autonomously generate text, including new creative content.

Yet, do human readers find such algorithm-generated text as appealing as – or more appealing than – human-written creative text? Do people's preferences differ when they are aware (transparency) versus unaware (opacity) of the algorithmic origin of the text? We experimentally examine how information about algorithms shapes humans' behavioral reactions, reflecting current directions in AI-safety research that deal with algorithmic transparency (Craglia et al., 2018; Garfinkel, Matthews, Shapiro, & Smith, 2017; Marcinkowski, Kieslich, Starke, & Lünich, 2020; Shin & Park, 2019).

Although transparency can refer to different types of disclosures around algorithmic decisions, here we focus on *algorithmic presence*, and hence the disclosure about whether an algorithm is involved in the decision (Diakopoulos, 2016). Transparency around algorithmic presence pertains to the current policy debate around whether people have a right to know when they are dealing with an algorithmic counterpart. For example, a proposed "Turing's red flag law" (Walsh, 2016) states, "An autonomous system should be designed so that it is unlikely to be mistaken for anything besides an autonomous system, and should identify itself at the start of any interaction with another agent" (Walsh, 2016, p. 35). Requests for such transparent information regimes have become increasingly voiced in light of the recently published hyper-realistic phone-call assistant Google Duplex (Leviathan & Matias, 2018) and robust text-generation algorithms such as GPT-2 (*The Guardian*, 2019).

When people are informed about algorithmic presence, extensive research reveals they are generally averse toward algorithmic decision-makers. This reluctance of "human decision makers to use superior but imperfect algorithms" (Burton, Stein, & Jensen, 2019; p.1) has been referred to as algorithm aversion (Dietvorst, Simmons, & Massey, 2015). In part driven by the belief that human errors are random, whereas algorithmic errors are systematic (Highhouse, 2008), people have shown resistance to algorithms in various domains (for a systematic literature review, see Burton et al., 2019). For example, people dislike, machines making moral decisions (Bigman & Gray, 2018), especially when they appear eerily human (Laakasuo, Palomäki, & Köbis, 2020), devalue purely algorithmic political choices (Starke & Lünich, 2020), and are even reluctant to rely on superior algorithmic recommendations about which jokes others would find funny (Yeomans, Shah, Mullainathan, & Kleinberg, 2019).

Regarding aversion to algorithm-generated text, research within digital journalism has assessed people's perceptions of news generated by algorithms (Carlson, 2015; Diakopoulos & Koliska, 2017). For example, companies such as *Automated Insights* produce articles for the Associated Press in domains where information exists in standardized formats, such as finance, sports, or weather. Experiments have compared people's evaluations of such algorithm-generated news pieces with those written by journalists (Clerwall, 2014; Graefe, Haim, Haarmann, & Brosius, 2018; Sundar & Nass, 2001). In one study, participants judged, among other facets, the overall quality, credibility, and objectivity of the text. The results reveal that algorithm-generated content is rated as more descriptive and boring, while also being viewed as objective and not necessarily distinguishable from content written by journalists (Clerwall, 2014). Another online experiment assessing people's perception of news pieces systematically manipulated the articles' actual and declared source (Graefe et al., 2018). Assessing credibility, readability, and journalistic expertise of the stimuli revealed participants consistently favored the allegedly human-written articles. People also reveal some aversion to algorithm-generated news, and thus *non-fiction* text.

Yet, do they equally dislike algorithm-generated *fiction*, that is, creative text? And does the information disclosure influence revealed preferences? Gaining answers to these questions is relevant to understanding the advances in artificial creativity and gauging the potential impact algorithms might have for creative industries (Bakhshi et al., 2015). Understanding whether people like or dislike creative text written by an algorithm also provides insights into whether NLG algorithms could be used to deceive others into believing the creative text stems from a human. That is, if people find the current output of algorithms such as GPT-2 entirely unappealing, the potential for ethical harm is less imminent. If, however, people find human and AI-written creative text comparably appealing, the door would open for AI to be used to craft such text on humans' behalf.

To find out whether people are averse to algorithm-generated creative texts, we assessed people's revealed preference for algorithm-generated creative text. From pairs of poems – each time, one originated from an algorithm and the other from a human – participants picked one they liked more. Between subjects, we either disclosed the respective origin of the poem (transparency) or not (opacity). We differed the degree of proficiency on the side of the human writers – untrained novices in Study 1 and experts in Study 2 – and compared their performance with the state-of-the-art algorithm GPT-2. We tested the prediction that humans would prefer the human-written poem, in particular when they were informed about the origin of the poem. Moreover, in Study 2, we additionally assessed stated preferences of algorithm aversion, by asking people how they generally perceive algorithms that write creative text, and tested our prediction that stated and revealed preferences are positively correlated.

## 1.4. Human Selection in and out of the Loop

Moreover, the combination of understanding people's accuracy in detecting algorithm-generated text and their preference for such text enables new insights into the deceptive potential of such NLG algorithms. That is, if people cannot tell human-written from algorithm-generated text and do not systematically prefer the former, GPT-2 and other algorithms might indeed be used for new forms of plagiarism. One key feature to understanding the deceptive potential of such algorithms is the degree of autonomy the algorithms have. Someone using the algorithm to craft text on one's behalf can scan through the outputs – algorithms such as GPT-2 are capable of creating multiple samples of text in mere seconds – and select the one most suitable for a particular task. In fact, media coverage such as *The Economist* interviewing GPT-2 has been criticized for "cherry picking" only GPT-2's most coherent and funny replies to make it appear more capable than it actually is (Marcus, 2020). Such human editing reflects a selection process with humans-in-the-loop (HITL, Goldenfein, 2019). On the other end of the spectrum are unfiltered algorithmic outputs, such as many chatbots, tweetbots, and other automated text-generating algorithms. These algorithms act autonomously. The selection process occurs with humans-out-of-the-loop (HOTL).

Previous research suggests human involvement in algorithmic decision-making crucially shapes perceptions of identical outcomes (Starke & Lünich, 2020), and the degree of a machine's autonomy drives moral evaluations (Bigman, Waytz, Alterovitz, & Gray, 2019). Yet, the behavioral reactions to these different regulation regimes remains largely unknown, in particular in relation to the generation of creative text. To gauge the state of the art in NLG creative-text generation, a human in the selection process plays a vital role. Therefore, we introduce an HITL and an HOTL treatment to the experiment to gauge the gain insights into how human involvement shapes behavioral reactions to the algorithm's performance. We tested the prediction that people's detection accuracy and revealed algorithm aversion of algorithm-generated poetry would drop when the poems were pre-selected by humans (vs. randomly sampled) from the outputs generated by GPT-2.

## 2.  General Method

In the set of pre-registered studies, we introduce established tournament designs from behavioral economics to computational creativity research, by creating a competition between two agents, and have an independent third party function as a judge (for similar set-ups, see Gneezy, Saccardo, & van Veldhuizen, 2019). Extending previous behavioral research, in which two humans competed with each other (for an overview, see Camerer, 2011), in our experimental set-up, humans directly compete with an AI agent, in this case, the NLG algorithm GPT-2.

Both studies entailed four parts (for an overview, see Table 1). Part 1 consisted of creating pairs of human-AI poems. On the human side, in Study 1, poems were written by participants who took part in an incentivized real-effort creative-writing task; in Study 2, we used existing professionally written poems. On the algorithm's side, the poem stemmed from a state-of-the-art NLG algorithm GPT-2. In Study 1, we, the authors, selected output that the algorithm generated. In Study 2, we introduced a between-subjects manipulation of selection procedure, namely, whether the poems entering the competition were again selected by the authors, that is, HITL, versus randomly sampled from the outputs that GPT-2 produced, that is, HOTL.

Part 2 entailed a judgment task. In it, a separate sample of participants acted as third-party judges and indicated their preference for the creative texts. In both studies, we manipulated between subjects whether participants received information about the origin of the text, that is, which of the two poems was written by a human. Comparing the Transparency treatment, in which participants were informed about the origin, with the Opacity treatment, in which they were oblivious, enables us to gain causal insights into how the information about algorithmic presence shapes revealed preferences. In Study 2, the selection manipulation of the HITL versus HOTL treatment additionally allowed us to test how human involvement in the selection process of the outputs of GPT-2 shapes these preferences.

Part 3 consisted of a new, incentivized version of the classical Turing Test (Saygin et al., 2012; Turing, 1950) to assess people's accuracy in identifying algorithm-generated creative text. Judges naïve to the origin of the poems faced the task of correctly distinguishing human-written from algorithm-generated text and stood to gain financially when they accurately did so. In Study 1, participants in the Opacity treatment engaged in this version of the Turing test, whereas in Study 2, we recruited a separate sample of participants. Study 2 further contained the selection treatment to assess how human involvement in the selection procedure (HITL vs. HOTL) shapes people's ability to differentiate human-written from algorithm-generated creative text.

In part 4, accompanying the accuracy assessment, participants indicated their confidence in identifying the correct poem. In Study 1, this measurement was unincentivized, whereas in Study 2, we attached financial incentives for judges correctly estimating their performance. Namely, they received a reward of €0.50 if they correctly indicated the number of rounds in which they identified the correct origin of the poem, which allows us to gauge how people's estimated performance compares with their actual performance, and how incentives influence a potential gap between the two.

## 2.1. Pre-registration statement

All studies reported in this manuscript are pre-registered on the Open Science Framework,[1] where we provide an overview of all hypotheses, pre-analysis plans, material, data, and R analysis scripts. We further provide several accompanying documents that provide background information and technical details on the use of the NLG algorithm, the procedure employed to gather and select the poems for the competitions.

---

[1] Pre-registration Study 1: https://osf.io/znjex

Pre-registrations for both parts of Study 2: https://osf.io/z6fhr & https://osf.io/uvmjx

**Table 1.** Overview of two studies that each contain four parts

|  | Study 1 | Study 2 |
|---|---|---|
| **Part 1 – Selection of poems as stimulus material** | Poems written by untrained writers ($N$=30)<br><br>vs.<br><br>GPT-2 Medium (final poems selected with HITL) | Professional poems (e.g., Maya Angelou)<br><br>vs.<br><br>GPT-2 XL (between-subjects treatment of final poems selected either with HITL or HOTL) |
| **Part 2 – Preference** | Participants ($N$=200) reveal preference for human-written vs. AI-generated poems while knowing the origin of the poems (Transparency) or not (Opacity) | Participants ($N$=400) reveal preference for human-written vs AI-generated poems while knowing the origin of the poems (Transparency) or not (Opacity) |
| **Part 3 – Detection Accuracy** | Incentivized version of Turing Test among participants in Opacity treatment ($N$ =100, reward = €0.50) | Incentivized version of Turing Test with separate sample ($N$ = 200, reward = €0.50) |
| **Part 4 – Confidence** | Unincentivized assessment of confidence in detection ability | Incentivized assessment of confidence of detection ability |

# 3. Study 1

## 3.1. Method

### 3.1.1. Part 1 – Selection of poems

**3.1.1.1. Participants and Procedure.** Thirty participants ($M_{Age}$ = 29.40, $SD_{Age}$ = 8.75; female = 56.67%) completed the task to write a poem and answer a few exit questions, which in total took, on average, around 11 minutes. To obtain high-quality online data, we recruited the participants via the online research platform Prolific Academic (for a discussion of different online research platforms, see Peer, Brandimarte, Samat, & Acquisti, 2017), paid participants an average of €15 per hour, and restricted the sample those who were proficient in English. After providing informed consent, participants were informed about the incentivized competition that they would enter (for full instructions, see SOM). Namely, they could win a prize of €2 if their text was chosen as the winner in the competition, which led to a total of €40 in bonuses being paid out.

**3.1.1.2. Human Competitor.** To enter the competition, participants had to write a short piece of poetry for which they received the first two lines. Participants could freely decide how to continue the poem, which had to be at least eight lines long and be written in English. Instructions further explained to participants that they should abstain from (a) writing gibberish (e.g., kajsdkjasdkjaskjd), (b) addressing the judge directly (e.g. "choose me as the winner"), and (c) plagiarizing other people's work, because doing any of the three would result in exclusion from the competition. Three independent naïve coders screened the entries according to whether the written texts adhered to these criteria. We randomly picked 20 poems that fulfilled the pre-specified criteria. The instructions explained this procedure to the participants, who, under the assumption that all participants fulfilled the inclusion criteria, had a 67% chance of being entered into the writing competition.

**3.1.1.3. AI Competitor.** The randomly picked poems written by participants were entered into a competition with poems written by GPT-2. Namely, for Study 1, we used the 345M model of GPT-2, which is the second model that OpenAI released (for the code, see https://github.com/openai/gpt-2). We trained GPT-2 for this specific writing task on a self-compiled dataset consisting of works from various professional poets, including Jane Campion, Roald Dahl, Robert Frost, and William Blake. Being trained to generate poems that abided by the above-mentioned rules, GPT-2 received the same two starting lines. To imitate poetry, text generated by GPT-2 further had to adhere to one of the pre-specified *poetry appearance* criteria to enter the writing competition. Namely, the poem had to use lines and verses, rhyme (ending or beginning rhyme, assonance), alliteration (words beginning with the same letter), onomatopoeia (phonetically imitating the sound of its meaning), rhythm, repetition, symbolism, or contradictions. Adhering to recent suggestions to increase reproducibility in AI research (Gibney, 2020), all the material including the training dataset is available on OSF. Of the GPT-2-written poems that qualified, 10 were chosen for the competition and randomly matched with a human-written poem.

### *3.1.2. Part 2-4 – Preference, Detection Accuracy, and Confidence*

**3.1.2.1. Participants.** Applying the same pre-selection criteria as in Stage 1, we recruited 200 participants via Prolific. Participants completed the task within 17 minutes, on average, yielding a mean payoff of €11.67 per hour. Applying the pre-registered attention check, we excluded eight participants. All results reported in this manuscript remaining qualitatively unaffected by the exclusions. The final sample consisted of 192 participants ($M_{Age}$ = 29.06, $SD_{Age}$ = 10.61; female = 39.1%, other/prefer not to indicate = 1%).

    **3.1.2.2. Preference.** Participants assigned to the role of judges received 10 pairs of poems, all beginning with the same two opening lines. For each pair, they had to indicate which poem they liked better, which had financial consequences because one of the 10 comparisons would be paid out, and the chosen winner would receive a prize of €2. Judges knew that in each pair, one poem stemmed from a human writer and the other poem stemmed from the algorithm GPT-2. To avoid efficiency concerns of judges seeking to reward the human writer, randomly chosen token players received the prize when the GPT-2-generated poem won the competition. That is, all participants knew that if GPT-2 won the pay-off-relevant round, another randomly selected writer received the reward.

    **3.1.2.3. Information Treatment**. Participants were randomly assigned to either the Transparency or the Opacity treatment. In the Transparency treatment, the judges received information about the origin of the poems. Hence, for each pair, they knew which poem was written by a human and which was generated by GPT-2, prior to their decision to award a winner. In the Opacity treatment, judges did not receive this information about the respective origin. Although knowing the pair of poems consisted of a human and an AI poem, judges did not know which was which.

    **3.1.2.4. Detection Accuracy.** Participants in the Opacity treatment, who were thus naïve about the origin of the poem, additionally engaged in an incentivized version of the Turing Test (Saygin et al., 2012; Turing, 1950). Akin to the original version proposed by Turing, judges faced the task of correctly distinguishing human- from machine-written text. In contrast to the standard version, however, judges could not directly interact with the two participants by asking

questions but merely received the text output. As a second refinement, we introduced incentives for accuracy. That is, judges could earn €0,50 if they correctly identified the origin of the text. They received 10 pairs of poems from which we randomly selected one for payment.

**3.1.2.5. Confidence.** Judges also estimated their level of confidence in correctly identifying the human poem on a 100-point scale (0 = *not at all confident*, 100 = *very confident*). We compared these subjective ratings of participants' confidence with the actual level of accuracy in determining the origin of the text.

## 3.2. Results

### 3.2.1. Part 1 – Selection of poems

Human-written and GPT2-generated poems did not significantly differ in length, as a signed-rank test on the number of words reveals ($p = .824$). Participants wrote a median of 37 words ($SD = 14.36$) using eight lines (7.55), whereas GPT-2 generated a median of 40 words ($SD = 12.94$), also using eight lines (7.75) on average. Thus, the poems were of similar length and could also not be distinguished visually or by other aesthetic rules. As outlined in the pre-registration, we collected additional exploratory variables assessing the human writers' level of confidence in winning the competition and the creativity ratings of the poems (see OSF).

### 3.2.2. Part 2 – Preference

Overall, human-written poems won 1,091 out of 1,915 competitions corresponding to a win share of 56.97%, which differs significantly from a win-share of 50% ($\chi^2 = 37.23$, $p < .001$). Mixed-effects probit regressions with random effects to account for dependencies of responses of individuals and per poem equally consistently reveal significant preferences for human-written over algorithm-generated poems as the significant intercept in Models 0-3 indicates (see

Table 2).[2] Hence, overall, judges showed a preference for human-written over algorithm-

generated poems.

---

[2] All mixed-effect regressions reported in the manuscript include random effects for the participant ID and the poem pair.

**Table 2**. Mixed-effects probit regressions predicting preference for the human-written poem in each round

|  | Model 0 | Model 1 | Model 2 | Model 3 |
|---|---|---|---|---|
| DV: Preference for human-written poetry |  |  |  |  |
| **(Intercept)** | 0.18** | 0.20** | 0.22** | 0.35* |
|  | (0.03) | (0.04) | (0.05) | (0.16) |
| **Treatment** |  | -0.06 | -0.06 | 0.07 |
|  |  | (0.06) | (0.06) | (0.06) |
| **Age** |  |  | -0.01 | -0.01 |
|  |  |  | (0.03) | (0.03) |
| **Gender** |  |  | -0.03 | -0.04 |
|  |  |  | (0.06) | (0.06) |
| **Education** |  |  |  |  |
| Primary School |  |  |  | -0.34 |
|  |  |  |  | (0.30) |
| High School |  |  |  | 0.04 |
|  |  |  |  | (0.07) |
| Master |  |  |  | 0.12 |
|  |  |  |  | (0.08) |
| PhD |  |  |  | 0.09 |
|  |  |  |  | (0.16) |
| **English Proficiency** |  |  |  |  |
| None |  |  |  | 0.04 |
|  |  |  |  | (0.29) |
| Limited Working |  |  |  | -0.15 |
|  |  |  |  | (0.18) |
| Professional Working |  |  |  | -0.17 |
|  |  |  |  | (0.16) |
| Full Professional |  |  |  | -0.15 |
|  |  |  |  | (0.17) |
| Native or Bilingual |  |  |  | -0.19 |
|  |  |  |  | (0.17) |
| *N* | 1915 | 1915 | 1905 | 1905 |

**Note.** Random effects included for the participants' ID and the pair of poems. Standard errors are reported in parentheses. DV = Preference, binary variable across 10 rounds coded as 0 = preference for algorithm-generated poem, 1 = preference for human-written poem. Independent variables: Age (continuous, standardized), Gender (dummy, reference category = male), Education (dummy, reference category = Bachelor's), Language: The Interagency Language Roundtable scale is used to determine the participants' level of English, with the reference category being elementary proficiency. Significance coding: * $p < .05$, ** $p < .01$, *** $p < .001$.

As a first test of whether people are more averse to algorithm-generated poetry in the Transparency than in the Opacity treatment, a two-sample t-test with equal variances on the number of wins for the human writers indicates no significant differences ($t(189) = 1.05$, $p = .29$, see also Figure 1).[3] Human writers won only slightly more often in the Transparency ($M = 5.82$, $SD = 1.69$) than in the Opacity treatment ($M = 5.59$, $SD = 1.49$). Because the assumption of normality of the aggregated wins was violated (Shapiro-Wilk: $W = 0.96$, $p < .001$), we conducted a Wilcoxon signed-rank test that similarly indicates no significant treatment differences ($mdn_{Opacity} = mdn_{Transparency} = 6$, $W = 4285$, $p = .39$).[4] Further, Bayesian independent-sample t-tests reveals a Bayes factor of $BF_{0+} = 7.40$, thus providing moderate support that the $H_0$ of no treatment differences is more likely than the $H_1$ of stronger preferences toward human-written poetry in the *Transparency* treatment. Finally, mixed-effects probit regressions predicting the binary outcome of preference in each round reveal no significant treatment differences, also when controlling for demographics and education levels (see Models 1-3, in Table 2). Taken together, these results suggest that, contrary to our hypothesis, judges did not reveal a stronger preference for human-written poetry when they were informed about the origin of the poems.

---

[3] Conservative sensitivity analysis with $\alpha = .05$, power of $1-\beta = .8$ and 100 participants per treatment suggests our analysis was able to detect a small effect (Cohen's $d = 0.39$).

[4] We report non-parametric tests throughout the manuscript when assumptions of normality were violated.

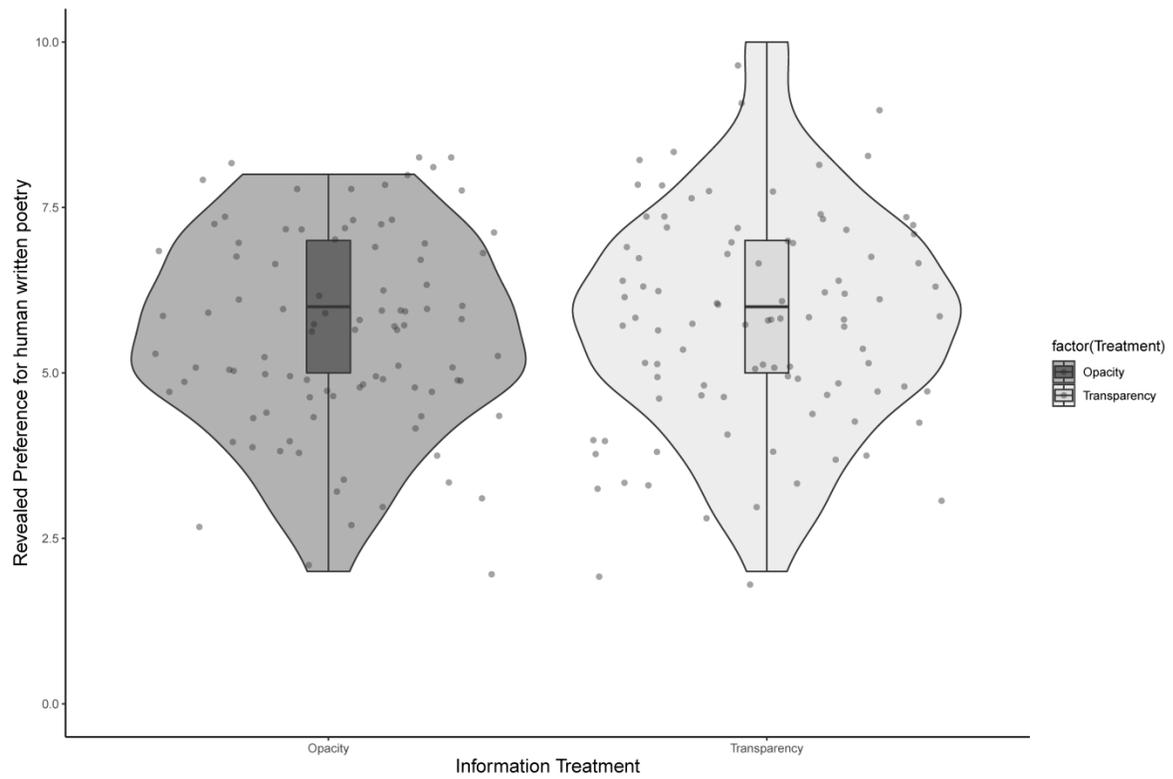

**Figure 1**. Violin plots of revealed preference for human-written poetry across information treatments.

**Note.** The plot depicts the sum score of rounds in which the participants chose the human-written poem in the Transparency treatment (left pane) and in the Opacity treatment (right pane). Inside the violin plot, mean and 95% confidence intervals are plotted, indicating a significant preference for human-written poems in both treatments yet no differences across treatments.

### 3.2.3. Part 3 – Detection Accuracy

Overall, judges identified the correct origin with an average accuracy of 50.21% (95%CI[46.4; 53.9]), which, according to a Wilcoxon signed-rank test, indicates no significant deviance from chance ($V = 1479$, $p = .935$).[5] A Bayesian binomial test yields a Bayes Factor of $BF_{01} = 24.91$, which strongly supports that the $H_0$ of judges' accuracy not exceeding chance is more likely than the $H_1$. Mixed-effects probit regressions, predicting the judges' accuracy in each round, reveal no significant deviation from chance at detecting the correct poem, also when controlling for standard demographics of age, gender, and education (see Table 3). Taken together, the results indicate people are not reliably able to identify human versus algorithmic creative content.

---

[5] Conservative sensitivity analysis with $\alpha = .05$, power of $1-\beta = .8$ and 100 participants and a normal parent distribution suggests our analysis was able to detect a small effect (Cohen's $d = 0.28$).

**Table 3.** Mixed-effects probit regressions on detection accuracy for each round

|  | **Model 0** | **Model 1** | **Model 2** |
|---|---|---|---|
| **DV: Detection Accuracy** | | | |
| **(Intercept)** | 0.01 | 0.01 | -0.35 |
|  | (0.05) | (0.14) | (0.27) |
| **Age** |  | 0.02 | 0.05 |
|  |  | (0.05) | (0.06) |
| **Gender** |  | -0.13 | -0.19 |
|  |  | (0.10) | (0.10) |
| **Education** | | | |
| High School |  |  | -0.04 |
|  |  |  | (0.11) |
| Master |  |  | -0.01 |
|  |  |  | (0.15) |
| PhD |  |  | -0.22 |
|  |  |  | (0.21) |
| **English Proficiency** | | | |
| None |  |  | 1.72* |
|  |  |  | (0.65) |
| Limited |  |  | 0.45 |
|  |  |  | (0.31) |
| Professional Working |  |  | 0.62* |
|  |  |  | (0.29) |
| Full professional |  |  | 0.33 |
|  |  |  | (0.28) |
| Native or bilingual |  |  | 0.47 |
|  |  |  | (0.28) |
| *N* | **733** | **733** | **733** |

**Note.** Random effects included for the participants' ID and the pair of poems. Standard errors are reported in parentheses. DV = Detection Accuracy, binary variable across 10 rounds coded as 0 = incorrect guess, 1 = accurate guess. Independent variables: Age (continuous, standardized), Gender (dummy, reference category = male), Education (dummy, reference category = Bachelor), Language: The

Interagency Language Roundtable scale is used to determine the participants' level of English, with the reference category being elementary proficiency. Significance coding: * $p < .05$, ** $p < .01$, *** $p < .001$.

### 3.2.4. Part 4 – Confidence

As non-pre-registered exploratory analyses, we examined judges' level of confidence in detecting the correct origin, prior to having read any samples. On a scale from 0 to 100, the average confidence level of the judges was $M = 62.27$ ($SD = 22.27$), which significantly exceeds chance levels ($t(732) = 14.92$, $p < .0001$). The distribution of confidence ratings is moderately left skewed (*skewness* = -0.496, $SE = 0.09$; see Figure 2, left pane). Hence, on aggregate, people rate their confidence level higher than chance. Regression analysis of peoples' confidence in differentiating human from GPT-2-written poems on their actual performance reveals no significant relationship ($b < .01$; $\beta = .017$, $t(74) = 0.143$, $p = .887$). Hence, self-rated confidence did not predict their actual performance. Moreover, we find a significant proportion of participants (69.33%) reveals overconfidence, defined as confidence levels exceeding participants' actual accuracy in their performance (see also Figure 2, right pane). Overall, these results provide a first tentative indication that people are not able to accurately predict, and instead overestimate, how well they would perform in the incentivized Turing Test.

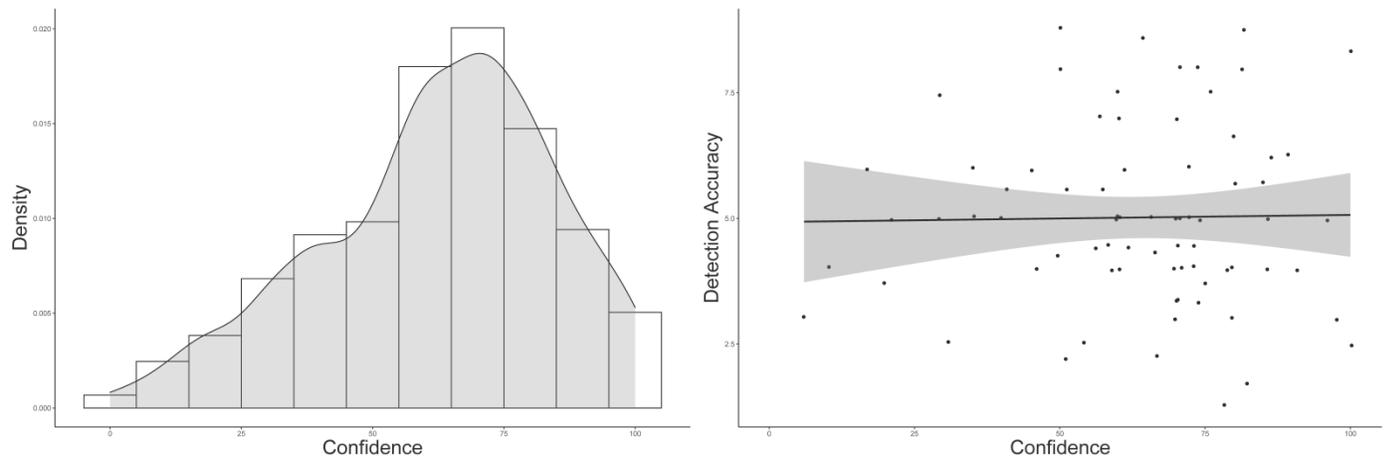

**Figure 2.** Density distribution of the judges' confidence score ranging from 0 to 100% (left pane). Scatterplot illustrating the relationship between confidence and their actual standardized performance across all rounds of the incentivized version of the Turing Test (right pane).

**3.2. Discussion**

Study 1 examined the behavioral responses to algorithm-generated creative text. The results reveal judges slightly preferred human-written to algorithm-generated poems, independent of whether they were cognizant (transparency) or oblivious (opacity) about the origin of the poem. This effect occurred even though their decisions had financial consequences for the writers. Moreover, in line with our expectations, judges were unable to reliably distinguish human from artificial poetry. In light of the financial rewards for accuracy in our version of the Turing Test, these results are among the first to indicate that detecting artificial text is not a matter of incentives but ability. At the same time, most judges' confidence levels exceeded their actual performance in recognizing artificial poetry – a first sign of overconfidence in algorithm detection.

## 4. Study 2

One potential criticism of Study 1 is that the comparison favored the algorithm. That is, poems created by novices competed with the output of a state-of-the-art algorithm trained on the works of prolific poets. Indeed, the recruited participants did not have any prior training in poetry writing and were put on the spot to write a poem within a short time frame. Moreover, the criteria for GPT-2-generated poetry were slightly stricter than for the human-written poetry, because GPT-2-generated poems had to adhere to an extra criterion of *poetry appearance* (e.g., using rhymes). Although algorithm-generated poetry was able to pass as written by a human, and people's preferences did not change according to whether they were informed about the origin of the poem, what happens when trained, or even professional poets, compete against an NLG algorithm remains unknown. To address these questions, we drew on existing poems written by renowned professional poets, such as Maya Angelou and Hermann Hesse, and entered them in the competition on the human side. On the algorithm's side, we fed the full model of GPT-2 the first two lines of these professional poems to generate samples of poems.

Further, to address the concern that GPT-2-generated poetry might have performed well due to stricter inclusion criteria, we introduced a new treatment, differing in the degree to which humans were involved in the selection of the generated poems. Ample AI research points out that whether humans are involved in the decision chain has a crucial impact on the performance of the algorithm and on people's evaluation of these outcomes (Rahwan, 2018; Schirner, Erdogmus, Chowdhury, & Padir, 2013; Starke & Lünich, 2020; Wang, Harper, & Zhu, 2020; Zhu, Yu, Halfaker, & Terveen, 2018). Yet, behavioral reactions to different levels of human involvement in algorithmic decision-making remain largely unknown. Studying machine and human behavior means dealing with autonomous, unpredictable outcomes on the sides of the human *and* the machine (Rahwan et al., 2019). In our case, the NLG algorithm GPT-2 produces samples of different outcomes each time it is run. For the purpose of the current study, the unpredictable outcomes raise the important question of how to determine which of the various outputs, namely, poems, that GPT-2 produces in a single run to use for the competition.

Two main strategies can be applied. The first is human selection, where humans pre-screen and select the poem they deem most suitable, that is, HITL. This strategy reflects the situation in which someone uses GPT-2 as a writing aid and selects the output deemed most useful. The second is random selection, in which the algorithmic output is randomly sampled and hence enters the competition unfiltered, with HOTL. This selection procedure reflects the unfiltered use of NLG algorithms such as for most tweet- or chatbots. Using HITL versus HOTL in the selection of poems this way, Study 2 provides some of the first answers to whether and how much these different strategies for the selection of the algorithm-generated content affect people's behavioral reactions. We thus again examined people's preferences, their accuracy in detecting algorithm-generated text, and their confidence levels of doing so.

**4.1. Method**

### 4.1.1. Part 1 - Selection of poems

**4.1.1.1. Human and AI Competitor.** We again created pairs of human-written and AI-generated poems. As outlined in more detail in the documentation of the stimulus material (see helper file on OSF), the human-written poems stem from a collection of poems written by professional poets. The AI-generated poems were generated using the full model of GPT-2, fed with two starting lines of the respective poem it was competing against as a prompt to generate a poem.

**4.1.1.2. Selection Treatment.** As a new treatment, we manipulated the way in which the algorithm-generated poem was selected. When letting GPT-2 generate text outputs, it produces samples of multiple poems at once. As a between-subjects manipulation, we differed how the poem was selected from this sample. In the HITL treatment, the authors (NCK & LDM) selected the best poem by consensus voting from the outputs generated by GPT-2. In the HOTL treatment, the poem was randomly sampled from the same outputs.

### 4.1.2. Part 2 – Preference

**4.1.2.1. Participants and Procedure.** For the poetry-judgment task, we recruited a sample of 400 participants via Prolific, paying on average €1.98 for a study that took around 16 minutes (= €7.43/hr). After applying the pre-registered exclusion criteria of excluding participants who failed the attention check, the final sample consisted of 384 judges ($M_{age}$ =

31.38, $SD_{age}$ = 11.92, female = 47.14%, other/prefer not to say = 0.54%). Judges read 10 poem pairs and, for each pair, picked the poem they liked more.

**4.1.2.2. Information Treatment.** Identical to Study 1, we again manipulated whether

participants were informed about the origin of the poem. Hence, in the Transparency treatment

($N$ = 192), judges knew which poem was written by a human and which was generated by AI,

whereas in the Opacity treatment ($N$ = 192), they did not.

**4.1.2.3. Algorithm-Aversion Scale.** To assess stated aversion to algorithmic poetry, we

included a new item to an existing scale to measure algorithm aversion (Castelo, Bos, &

Lehmann, 2019). The scale consists of multiple items, each describing different tasks (e.g.,

"driving a car"), for which participants have to indicate whom they trust more to execute that

task. Answers are given on a 100-point slider scale ranging from 0 (=a qualified human) to 100

(=an algorithm). To the list of existing items, we added the new item, "Writing poetry."

### 4.1.3. Part 3 & 4 – Detection accuracy & confidence

**4.1.3.1. Participants & Procedure.** To assess detection accuracy and confidence in

detecting algorithm-generated text, we recruited a separate sample of 200 participants via

Prolific for a study that took, on average, 13.92 minutes and paid €2.26 (= €9.74/hr). After

applying the pre-registered exclusion criteria, the final sample consisted of 185 participants

($M_{Age}$= 27.66; $SD_{Age}$ = 9.47, female = 47.02%).

**4.1.3.2. Detection accuracy.** Identical to Study 1, we used the incentivized version of the

Turing Test in which people could receive a financial reward of €0.50 for correctly identifying the origin of a poem, that is, whether it was human-written versus algorithm-generated.

**4.1.3.3. Confidence.** After participants completed the accuracy assessment, they were asked to estimate in how many of the rounds they correctly identified the origin of the poem. We incentivized this elicitation of confidence by rewarding the correct estimation of the number of rounds with a financial bonus of €0.50. We assessed the estimated performance after and not before participants completed the incentivized Turing Test to avoid hedging, namely, participants changing their performance in the task to match their estimation.

**4.1.3.4. Knowledge of Poetry.** After participants completed Parts 2-4, we assessed their prior poetry knowledge. We presented the poems used in the study and asked two questions. First, as a *stated* poetry-knowledge assessment, we asked the participants whether they had read the poem prior to participating in this study (Y/N). Second, as a measure of *revealed* poetry knowledge, we asked them to impute the respective poet's name.

**4.1.3.5. Demographics.** Using the same questions as in Study 1, we again assessed all participants' standard demographic information of age, gender, and education, as well as their experience with computer science, and their views on the development of general artificial intelligence at the end of the study.

## 4.2. Results

### 4.2.1. Part 2 – Preference

Testing whether judges overall preferred the human-written over the AI-generated poems, a $\chi^2$ test comparing the observed human win share with a chance-level win share of 50% suggests a significant deviation ($\chi^2(1) = 340.82$, $p < .001$). Human writers overall won 64.90% of the comparisons. A t-test similarly reveals human-written poems were chosen more often across the 10 rounds ($M = 6.49$, $SD = 1.65$) than would be expected by chance ($t(383) = 17.73$, $p <$

.001). Mixed-effects probit regressions reveal significant intercepts in all models (see Table 4), indicating a preference for human-written poems. Taken together, the results replicate results obtained in Study 1 and confirm our hypothesis that people overall reveal preferences for human-written poems over algorithm-generated poems.

Examining whether judges reveal a stronger preference for human-written poems in the *Transparency* (vs. *Opacity*) treatment, a two-sample t-test comparing the mean number of human wins reveals no significant treatment differences ($t(365) = 0.62$, $p = .54$). In fact, judges selected human poems slightly less often when they knew the origin ($M = 6.44$, $SD = 1.46$) than when they did not ($M = 6.54$, $SD = 1.82$). Similarly, non-parametric tests reveal no differences (WSR: $W = 17961$, $p = 0.44$, $mdn_{Transparecny} = 6$, $mdn_{Opacity} = 7$), and Bayesian analyses provide strong evidence that no treatment differences are more likely than the expected treatment differences ($BF_{01} = 13.55$). Mixed-effects probit regressions predicting the binary preference measure in each round reveal no significant information treatment effects either (see Models 1,3 & 4 in Table 4). Hence, contrary to our hypothesis, people did not reveal stronger preferences for human-written poems in the Transparency treatment than in the Opacity treatment.

Testing whether judges revealed a stronger preference for human-written poems over AI-generated poems in the HOTL (vs. HITL) treatment, a t-test of the mean number of human wins suggests significant differences ($t(372) = -2.82$, $p = .005$). Judges selected the human poem on average more often in the HOTL ($M = 6.69$, $SD = 1.69$) than in the HITL treatment ($M = 6.23$, $SD = 1.55$). Similar results are obtained for non-parametric tests (WSR: $W = 15256$, $p = .007$; $mdn_{HOTL} = 7$, $mdn_{HITL} = 6$). Mixed-effects probit regressions predicting the preference each round also consistently reveal significant treatment differences (see selection treatment dummy in Models 2-4, Table 4). Taken together, the results suggest judges reveal a stronger preference for the human-written poems when the algorithm-generated poems are randomly sampled (HOTL) than when they are selected by humans (HITL).

**Table 4**. Mixed-effects probit regressions predicting preference for the human-written poem in each round

|  | **Model 0** | **Model 1** | **Model 2** | **Model 3** | **Model 4** |
|---|---|---|---|---|---|
| DV: Preference for human-written poetry |  |  |  |  |  |
| **(Intercept)** | 0.66*** | 0.69*** | 0.51*** | 0.35*** | 0.35*** |
|  | (0.15) | (0.15) | (0.06) | (0.10) | (0.10) |
| **Information Treatment** |  | -0.05 |  | -0.03 | -0.03 |
|  |  | (0.08) |  | (0.05) | (0.05) |
| **Selection Treatment** |  |  | 0.21*** | **0.13 | 0.14*** |
|  |  |  | (0.07) | (0.05) | (0.05) |
| **Revealed Poetry Knowledge** |  |  |  | -0.16 | -0.15 |
|  |  |  |  | (0.13) | (0.13) |
| **Age** |  |  |  |  | 0.08 |
|  |  |  |  |  | (0.02) |
| **Gender** |  |  |  |  |  |
| Male |  |  |  |  | -0.01 |
|  |  |  |  |  | (0.05) |
| Other |  |  |  |  | 0.60 |
|  |  |  |  |  | (0.36) |
| *N* | 3840 | 3840 | 3840 | 3840 | 3840 |

**Note.** Random effects included for the participants' ID and the pair of poems. Standard errors are reported in parentheses. DV = Preference, binary variable across 10 rounds coded as 0 = preference for algorithm-generated poem, 1 = preference for human-written poem. Independent variables: Information treatment (dummy, reference category = Opacity), Selection treatment (dummy, reference category = HITL), Revealed poetry knowledge (continuous), Age (continuous, standardized), Gender (dummy, reference category = female). Significance coding: * $p < .05$, ** $p < .01$, *** $p < .001$.

Besides *revealed* preferences for human-written poetry, we additionally analyzed people's *stated* preferences. Responses to the item assessing whether people prefer humans (0) or algorithms (100) to write poetry reveals an average score of $M = 19.50$, $SD = 20.73$ – a significant negative deviation from the mid-point of the scale ($t(383) = -28.83$, $p < .001$). This finding indicates people generally state that they prefer that humans as opposed to algorithms write poetry. Testing whether stated and revealed preferences for human versus algorithmic poetry are related, a point-biserial t-test between the stated preference item and the number of human wins reveals a significant positive correlation ($t(382) = 2.55$, $r = 0.13$, $p = .01$). Mixed-effects probit regressions similarly reveal a positive association ($b = 0.06$, $SE = 0.03$, $Z = 1.93$, $p = .05$). The marginal effect is $b = 0.06$, $SE = 0.02$, $Z = 2.55$, $p = .01$, which remains significant when controlling for both treatments, knowledge of the poets, and demographic information of gender and age (all $b$s > 0.05, $p$s < .03). Taken together, we find evidence for a weak but significant link between stated and revealed preferences in the domain of algorithmic poetry.

### 4.2.2. Part 3 – Detection accuracy

Testing whether people's poem-detection accuracy exceeds chance levels, a one-sample t-test comparing the aggregated accuracy across all rounds ($M = 5.94$, $SD = 2.01$) to a chance level of 5 reveals a significant difference ($t(184) = 6.33$, $d = 0.47$, $p < .001$). Non-parametric tests reveal similar results (Wilcoxon $V = 8406$, $p < .001$, $mdn_{accuracy} = 6$). Mixed-effects probit regressions predicting detection accuracy in each round reveal a significant intercept when not including control variables (see Model 0, in Table 4). In sum, these findings suggest that overall, participants were able to detect the correct origin of the poems at better than chance levels.

**Table 5.** Mixed-effects probit regression predicting the detection accuracy of the origin of the poem per round

|  | Model 0 | Model 1 | Model 2 | Model 3 |
|---|---|---|---|---|
| DV: Detection Accuracy | | | | |
| **(Intercept)** | 0.26*** | 0.10 | 0.09 | 0.20 |
|  | (0.09) | (0.09) | (0.09) | (0.10) |
| **Selection Treatment** |  | 0.34*** | 0.33*** | 0.32*** |
|  |  | (0.08) | (0.08) | (0.08) |
| **Revealed Poetry Knowledge** |  |  | 0.62** | 0.67** |
|  |  |  | (0.25) | (0.25) |
| **Age** |  |  |  | 0.00 |
|  |  |  |  | (0.04) |
| **Gender** |  |  |  | -0.19* |
|  |  |  |  | (0.08) |
| *N* | 1850 | 1850 | 1850 | 1850 |

**Note.** Random effects included for the participants' ID and the pair of poems. Standard errors are reported in parentheses. DV = Detection accuracy, binary variable across 10 rounds coded as 0 = incorrect guess, 1 = accurate guess. Independent variables: Selection treatment (dummy, reference category = *HITL*), Revealed poetry knowledge (continuous) Age (continuous, standardized), Gender (dummy, reference category = female). Significance coding: * $p < .05$, ** $p < .01$, *** $p < .001$.

As a first test for whether accuracy levels are higher in the HOTL than in the HITL treatment, a two-sample t-test comparing accurate guesses across all 10 rounds reveals a significant difference ($t(183)$= -4.19, $d$ = -0.62, $p < .001$). Participants identified the correct origin of the poems more frequently in the HOTL treatment ($M = 6.55$, $SD = 1.90$) than in the HITL treatment ($M = 5.37$, $SD = 1.95$). Similarly, results of non-parametric tests reveal significant differences (Mann-Whitney $U = 2872.5$, $r = -0.33$; $p < .001$; $mdn_{HOTL} = 7$, $mdn_{HITL} = 5$). Mixed-effect probit regressions also reveal significant treatment differences (see Models 1-3 in Table 5). Taken together, the results provide support for the predicted effect that people are better at detecting the origin of the poem for randomly chosen poems generated by GPT-2 (*HOTL*) than for human-selected poems (*HITL*).

Further, subgroup analyses reveal that whereas the accuracy rates in the HOTL treatment deviate significantly from chance (Student: $t(88) = 7.72$, $p < .001$; WSR: $V = 2304$, $p < .001$), in the HITL treatment, they do not (Student: $t(95) = 1.83$, $p = .07$, WSR: $V = 1820.5$, $p = .1$, see also Figure 3). To further corroborate these subgroup effects, we conducted non-pre-registered Bayesian analyses. The results reveal strong evidence that in the HOTL treatment, people's accuracy significantly positively deviated from chance ($BF_{10} = 5.095e+8$). By contrast, in the HITL treatment, the results provide anecdotal evidence in support of the null hypothesis that people are not better than chance at detecting the correct origin ($BF_{01} = 1.79$). Similarly, subgroup analysis using mixed-effects probit regressions suggest people are better than chance at detecting the HOTL selected poems ($b = 0.44$, $SE = 0.10$, $Z = 4.53$, $p < .001$), while not deviating significantly from chance in the HITL treatment ($b = 0.10$, $SE = 0.09$, $Z = 1.06$, $p = .26$). These patterns remain robust when introducing control variables of demographics and knowledge of the respective poem (see Models 2 & 3, Table 5). Taken together, these findings support the prediction that people's ability to detect the correct origin of a poem depends on the way the poems are selected. Although people can distinguish professional poems from algorithm-generated poems that are randomly chosen with an HITL, they cannot reliably do so when these poems are selected by an HOTL.

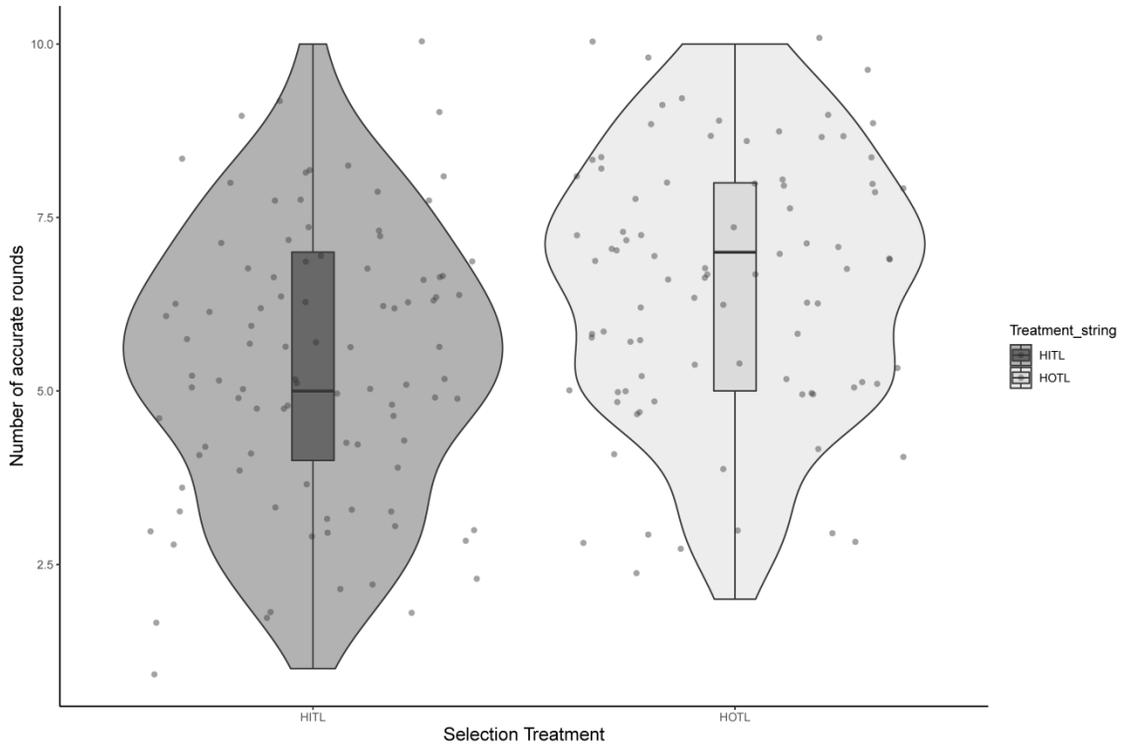

**Figure 3.** Violin plots depicting the distribution of accurate rounds across the selection treatment.

**Note.** The plot depicts the distribution of aggregated rounds in which the participants correctly identified the origin of the poem across in the HITL treatment (left pane) and HOTL treatment (right pane). Inside the violin plot, mean and 95% confidence intervals are plotted, indicating a significant ability of people to identify the correct origin only in the HOTL treatment preference for human-written poems in both treatments.

### 4.2.3. Part 4 – Confidence

As a first analysis of overconfidence, we examined the distribution of self-reported confidence ratings. On a 10-point scale, average confidence level of the judges was $M = 5.99$ ($SD = 1.77$), with the distribution being slightly left skewed (*skewness* = 0.26, $SE = 0.18$ see Figure 4, left pane). Using the same classification of overconfidence as in Study 1 (confidence - performance), 38.91% of the participants displayed overconfidence. A linear regression of confidence predicting accuracy levels to assess whether actual and believed performance are correlated indicate a significant, positive relationship ($b = 0.93$, $SE = 0.03$, $t(185) = 34.92$, $p < .0001$, see Figure 4, right pane).

Tests examining whether estimated accuracy levels significantly exceed actual accuracy levels reveal no significant differences ($t(184) = -0.35$, $p = .76$; WSR: $W = 17045$, $p = .95$). Mixed-effects linear regressions predicting overconfidence (i.e., confidence - accuracy) also reveal no significant intercept, also when controlling for demographics and stated as well as revealed poetry knowledge ($bs < 0.06$, $ps > .18$). In concert, these results suggest people overall show few signs of overconfidence and rather accurately estimate their ability to detect algorithm-generated poems.

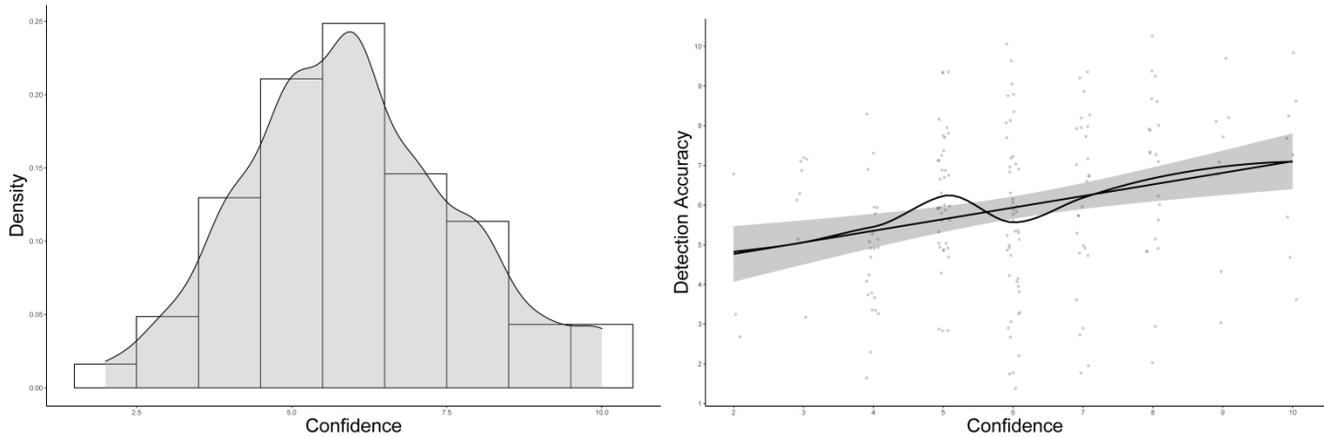

**Figure 4**. Density distribution of the judges' confidence score ranging from 0 to 10 (left pane). Scatterplot illustrating the relationship between people's estimated (x-axis) and actual (y-axis) detection accuracy (right pane). The graphs plot linear regression slope as well as a slope with binomial smoothened estimates.

### 4.3. Discussion

Study 2 replicated two main findings observed in Study 1 and provides novel insights indicating humans being in versus out of the loop in process of selecting the poems crucially shapes both preferences and detection accuracy. First, the findings again reveal people prefer human-written to algorithm-generated poems, which is unaffected by the information about algorithmic presence, and hence whether the origin of the poem was transparently communicated or remained opaque. As a second replication, people were again unable to reliably distinguish human from artificial poetry, while being incentivized to do so. However, this inability only occurred when humans could handpick the best poems (i.e. in the HITL treatment). When poems were randomly selected (i.e., in the HOTL treatment), people could detect the algorithm-generated poem with higher-than-chance levels. Lending further credence to the importance of the selection procedure involved, the results equally show significantly higher preference for algorithm-generated poems when humans were involved in the selection process. As some of the first insights into the behavioral responses to different levels of human involvement in the selection process of AI-generated content, the results show that humans being involved or not in the selection process strongly influences the abilities of the algorithm.

The results provide nuance to the link between estimated and actual algorithm-detection accuracy. Whereas in Study 1, participants' unincentivized estimated performance significantly exceeded their actual performance, Study 2 elicited these estimations in an incentivized way. Contrary to the results of Study 1 and the hypothesis, the estimated performance did not significantly exceed actual performance, and a positive link between detection accuracy and confidence therein existed.

## 5.   General Discussion

Algorithms increasingly influence humans' daily lives. Due to their growing learning abilities, autonomy, and unpredictability in outcomes, understanding such machine behavior, and how it affects human behavior, becomes vital (Rahwan et al., 2019). The current set of studies contributes to this research by examining behavioral responses to the state-of-the-art NLG algorithm, GPT-2. Our results provide four main insights. First, although participants stated and revealed aversion to artificial poetry, this aversive tendency did not increase when they were informed about the algorithmic origin of the text. These results bear special relevance when considering the second main insight: participants were incapable of reliably detecting algorithm-generated poetry, even when they were incentivized to do so and when the algorithm competed with esteemed poets. Third, although overconfidence in the algorithm-detection abilities existed when assessed in an unincentivized way (Study 1), no sign of systematic overconfidence existed when measured in an incentivized way (Study 2). Finally, the results of Study 2 point toward the important role that humans play in the implementation of algorithmic outputs: humans involved in the process of selecting poems reduce revealed algorithm aversion and detection accuracy. In fact, when people are not involved in the selection process, accuracy did exceed chance levels. We discuss the implications of each of these insights in turn.

## 5.1. Artificial Creativity: Aversion and Appreciation

Although first algorithms reach, and even surpass, human capacities in many narrow tasks, humans often show a general aversion to adopting algorithms (for a review, see Burton et al., 2019). In line with these findings, participants' views on algorithms crafting poetry were aversive, and these views correlated consistently, but weakly, with their behavior in choosing human poems over algorithmic poems. Reflecting current policy debates about transparency of algorithmic presence (Diakopoulos, 2016), our experiments examine the interplay of information and preferences for artificial versus human text outputs. Contrary to our expectations, participants revealed no stronger algorithm aversion when informed about the algorithmic origin of the text.

Our findings thus contribute to ongoing research seeking to disentangle when people are averse to (vs. appreciative of) algorithmic decision-making across various domains (Castelo et al., 2019; Lee, 2018; Pew Research Center, 2018). One key finding arising from that literature is that people dislike algorithms to execute emotional (vs. mechanical) tasks (Castelo et al., 2019). Hence, one interpretation of our results documenting aversion to algorithmic poetry is that people view writing poetry as an emotionally charged task. We derive first support for that notion from our data collected using the existing algorithm-aversion items. One of the original items asked participants about their views about algorithms writing newspaper articles. Comparing the views about algorithms performing these different language-generation tasks – writing poetry versus newspaper articles – indicates people are significantly more approving of algorithms in the role of journalists than in the role of poets (see full analysis in OSF).

## 5.2. Distinguishing between Artificial and Human

The question we brought to the online lab – whether people are able to distinguish artificial from human poetry – has attracted academic (Oliveira, 2009; Riedl, 2014) and public attention (Schwartz, 2015). For example, in a TED talk with more than 850,000 views, Oscar Schwartz compares poems by poets with generative poetry, and based on his results, claims computers can indeed write poetry (Schwartz, 2015). Here, we extend such previous approaches in two fundamental ways. First, instead of using generative poetry algorithms that are specifically developed to merely write poetry, we use GPT-2, an algorithm more robust to different environments. The fact that although the algorithm is not specifically tailored to generate poetry, yet still manages to pass as a human writer, underlines the purported abilities of the algorithm to create human-like text (Radford et al., 2019).

Second, we deviate from previous approaches by introducing financial incentives to a version of the Turing Test. Financially incentivizing choices is common in behavioral research, aimed at reducing measurement errors by increasing people's accuracy (Ariely & Norton, 2007). The results of both studies substantiate the view that differentiating between human-written and algorithm-generated poetry is not a matter of effort, but ability. Moreover, as we return to below, gaining a definite answer about whether a computer can write poetry that passes as human depends on whether and how humans are involved in the process of selecting the output.

## 5.3. Confidence in Algorithm-Detection Abilities

Incentives seem to also play a role in people's estimation of their own abilities in detecting algorithmic poetry. Although the majority of participants displayed overconfidence when confidence was elicited in an unincentivized way (Study 1), using incentivized measures of confidence produced no evidence for systematic overconfidence (Study 2). Instead, detection accuracy and confidence were positively correlated. When people stood to gain financially from detecting the origin of the poems, they seemed to calibrate their responses, because their estimated and actual performance overall matched well.

This lack of overconfidence is remarkable in light of participants' inexperience with the task. Participants were not able to draw on prior knowledge of their ability to detect algorithmic poetry yet were still able to provide informative estimates when reflecting on their own performance. Taken together, the results suggest overconfidence in algorithm detection can be curbed by providing financial rewards so that people strive to accurately estimate their own performance. Although the two studies also differ on other aspects – most notably the relative quality of the human and the AI poems – previous research corroborate the claim that incentives indeed lead to more precise confidence ratings (Schlag, Tremewan, & van der Weele, 2015).

## 5.4. Human Selection in and out of the Loop

Our results suggest NLG algorithms can generate poems that pass as human and that the poems are considerably appealing to readers, even when competing with the work of professional writers. However, the results of Study 2 suggest humans play an integral role in the process – only poems selected by the experimenters successfully passed as human and lowered algorithm aversion. Hence, whether humans are in or out of the selection loop shaped participants' reactions to the algorithm's performance.

Thereby, we provide some of the first behavioral insights into people's reactions to different HITL systems, and complement a rich (technical) literature in AI research (Schirner et al., 2013). Seeking to mitigate the limitations of algorithms, HITL systems have been proposed to increase algorithmic accountability (Wang et al., 2020; Zhu et al., 2018), because keeping humans in the loop helps to monitor and adjust the system and its outcomes (Rahwan, 2018). Here, we show that humans in the loop also allow us to harness the potential of recent developments in NLG, and crucially shape the conclusions drawn about the machine's behavior.

## 5.5. Implications and Future Research

The results of the studies have (ethical) implications. Language-generation algorithms are entering daily lives. Using transfer learning, GPT-2 can be fine-tuned to craft text in domains other than poetry, such as crafting artificial online reviews, patent claims (Lee & Hsiang, 2019), or fake tweets (Ressmeyer, Masling, & Liao, 2019), but also assist creative writers or provide useful feedback for customers (Budzianowski & Vulić, 2019). NLG algorithms thus have potential but also perils. To contribute to responsible use, future experimental studies examining how people react to algorithm-generated text in different domains will help provide valuable empirical insights.

Our experimental framework contributes to the methodological toolkit to systematically study the impact of NLG algorithms on human behavior. To gain insights into the question of whether people are able to detect algorithm-generated text, future studies using the novel incentivized version of the Turing Test could examine NLG's abilities in other domains. For example, studies on automated news generation (Carlson, 2015; Diakopoulos & Koliska, 2017) or longer AI-generated (creative) texts could unveil whether AI similarly can pass as human in these domains.

Studies seeking to investigate existing NLG algorithms face the challenge of unpredictable and changing text outputs, which leads to less experimental control over the machine's behavior (Rahwan et al., 2019), yet provides new researchers degrees of freedom in stimulus selection. Our new treatment comparing HITL with HOTL indicates this methodological choice influences the results. Future research could extend the external reliability of the current design by letting fellow participants be the ones who select the text output. We hope the current set-up encourages standardized selection protocols paired with open science practices (Srivastava, 2018) to gain reliable and reproducible findings on the nexus of human and machine behavior.

Taking a step back and examining the overall pattern of results, we emphasize that the results do not indicate machines are "creative." In fact, one of the main functions of creativity in general and in poetry in particular is the expression of (deep) emotions, a feat that machines lack (so far). The results are rather a testament to the increasing abilities of NLG algorithms to create text that mimics human creative text and that people do find appealing. Algorithms such as GPT-2 are widely assumed to have a long way to go before they can autonomously write truly creative text, especially in longer formats than poems. However, projects in which humans and algorithms form hybrid writing teams and collaboratively craft fiction text present one way in which such algorithms could enter our daily lives. Whether such forms of hybrid collaborations between human and machines should be considered plagiarism remains unclear. Or, conversely, to what extent does the (developer of the) algorithm deserve (financial) credit for the textual outputs? Related to the set-up used in the current studies, would an entry to an actual poetry competition by a contestant who uses GPT-2 input be counted as fraud? If so, how could it be detected? And if not, (how) should the prize money be split?

**5.6.Conclusion**

Algorithms that generate text resembling human language are becoming ever more widely accessible. Not only novelists with writer's block can make use of freely available algorithms like GPT-2. Understanding humans' behavioral reactions to such algorithms helps shape policies to ensure artificial intelligence remains beneficial (Crawford & Calo, 2016). As a step in that direction, the present set of studies adopts a behavioral science approach to examine creative artificial intelligence. We hope more studies follow suit to inform policies of disclosure of algorithmic presence and provide new behavioral insights into human versus (creative) artificial intelligence.